\documentclass[aps,pre,showpacs,amsmath,amssymb,
floatfix,nofootinbib]{revtex4}
\usepackage{graphicx} \usepackage{dcolumn} \usepackage{bm}
\usepackage{tipa} \usepackage{CJK} \usepackage{subfigure}
\newcommand{\comment}[1]{}
\renewcommand{\d}{{\rm d}}

\begin{document}
\draft


\title{Stochastic model for phonemes uncovers an author-dependency of
  their usage}

\author{ Weibing Deng$^{1)}$ and Armen E. Allahverdyan$^{2),}$
\footnote{wdeng@mail.ccnu.edu.cn, armen.allahverdyan@gmail.com}}

\address{ $^{1)}$Murray Gell-Mann Institute of Complexity Science,
  Central China Normal University, Wuhan 430079, China \\
  $^{2)}$Yerevan Physics Institute, Alikhanian Brothers Street 2,
  Yerevan 375036, Armenia}

\begin{abstract}

  We study rank-frequency relations for phonemes, the minimal units
  that still relate to linguistic meaning. We show that these
  relations can be described by the Dirichlet distribution, a direct
  analogue of the ideal-gas model in statistical mechanics. This
  description allows us to demonstrate that the rank-frequency
  relations for phonemes of a text do depend on its author. The
  author-dependency effect is not caused by the author's vocabulary
  (common words used in different texts), and is confirmed by several
  alternative means. This suggests that it can be directly related to
  phonemes. These features contrast to rank-frequency relations for
  words, which are both author and text independent and are governed
  by the Zipf's law.

\end{abstract}

\pacs{89.75.Fb, 05.10.Gg, 05.65.+b}


\maketitle

\section{Introduction}
\label{intro}


Language can be viewed as a hierarchic construction: phoneme,
syllable, morpheme, word ... Each of these objects expresses meaning
or participates in its formation, and consists of elements of the
previous level, i.e. syllable consists of phonemes
\cite{scherba,phoneme_def,sapir}.

The lowest hierarchic level is phoneme, which is defined to be a
representative for a group of sounds that are not distinguishable with
respect to their meaning-formation function in a concrete
language. For instance {\it /r/} and {\it /l/} are different phonemes
in English, e.g. because {\it row} and {\it low} which differ only by
these phonemes are different words; see section I of the supplementary
material for a list of English phonemes. But they are the same phoneme
in Japanese, since in that language there is no danger of
meaning-ambiguity upon mixing {\it /r/} with {\it /l/}.  (Different
speech sounds that are realizations of the same phoneme are known as
allophones.) Thus the meaning is crucial for the definition of the
phoneme, although a single phoneme does not express a separate meaning
\cite{scherba,phoneme_def,sapir}. The next hierarchic level (syllable)
indirectly participates in the definition of the phoneme, since the
syllable bounds phonemes, i.e. there cannot be a phoneme which belongs
to two different syllables; e.g.  diphthongs belong to the same syllable
\cite{scherba,phoneme_def}.

The history of phoneme is a rich and complex one. It appeared in Greek
and Indian linguistic traditions simultaneously with atomistic ideas
in natural philosophy \cite{skoyles,staal,lysenko}. Analogies between
atom and phoneme are still potent in describing complex systems
\cite{zwick,abler}. Within the Western linguistic tradition the
development of phoneme was for a while overshadowed by related (but
different) concepts of letter and sound
\cite{scherba,phoneme_def}. The modern definition of phoneme goes back
to late XIX century \cite{phoneme_def}. While it is agreed that the
phoneme is a unit of linguistic analysis \cite{sapir}, its
psychological status is a convoluted issue
\cite{port,suomi,nathan,savin,foss}. Different schools of phonology
and psychology argue differently about it, and there is a spectrum of
opinions concerning the issue (e.g. perception of phonemes, their
identification, reproduction {\it etc}) \cite{savin,foss}; see
\cite{port,suomi,nathan} for recent reviews.


For defining a rank-frequency relation, one calculates the frequencies
$f_r$ of certain constituents (e.g. words or phonemes) entering into a
given text, lists them in a decreasing order
\begin{eqnarray}
  \label{eq:0}
  f_1\geq f_2\geq ...\geq f_n,
\end{eqnarray}
and studies the dependence of the frequency $f_r$ on the rank $r$ (its
position in (\ref{eq:0}), $1\leq r\leq n$). This provides a
coarse-grained description, because not the frequencies of specific
phonemes are described, but rather the order relation between them,
e.g. the same form of the rank-frequency relation in two different texts
is consistent with the same phoneme having different frequencies in
those texts. The main point of employing rank-frequency relations is
that they (in contrast to the full set of frequencies) can be
described via simple statistical models with very few parameters.

Rank-frequency relations are well-known for words, where they comply
to the Zipf's law; see \cite{wyllis,baa} for reviews. This law is
universal in the sense that for all sufficiently long texts (and their
mixtures, i.e. corpora) it predicts the same power law shape
$f_r\propto r^{-1}$ for the dependence of the word frequency on its
rank. It was shown recently that the representation of the word
frequencies via hidden frequencies|the same idea as employed in the
present work|is capable of reproducing both the Zipf's law and its
generalizations to low-frequency words (hapax legomena)
\cite{zipf_pre}. Due to its universality, the Zipf's law for words
cannot relate the text to its author.

The rank-frequency relation for morphemes and syllables was so far not
studied systematically. Ref.~\cite{zipf_chin} comes close to this
potentially interesting problem, since it studies the rank-frequency
relations of Chinese characters, which are known to represent both
morpheme and syllable (in this context see also
\cite{am_j,sh,chen_guo}). This study demonstrated that the Zipf's law
still holds for a restricted range of ranks. For long texts this range
is relatively small, but the frequencies in this range are important,
since they carry out $\simeq 40$ \% of the overall text frequency. It
was argued that the characters in this range refer to the most
polysemic morphemes \cite{zipf_chin}.

There are also several works devoted to the rank-frequency relations
of phonemes and letters
\cite{sigurd,good,gusein_1,gusein_2,witten_bell,tambov,hindu}. One of
first works is that by Sigurd, who has shown that the phoneme
rank-frequency relations are not described by the Zipf's law
\cite{sigurd}. He also noted that a geometric distribution gives a
better fit than the Zipf's law.  Other works studied various
few-parameter functions|e.g. the Yule's distribution|and fitted it to
the rank-frequency relations for phonemes of various languages; see
\cite{hindu} for a recent review of that activity.

The present work has two motivations. First, we want to provide an
accurate description of rank-frequency relation for phonemes. It is
shown that such a description is provided by postulating that phoneme
frequencies are random variables with a given density. The ranked
frequencies are then recovered via the order statistics of this
density. This postulate allows to restrict the freedom of choosing
various (theoretical) forms of rank-frequency relations, since|as
developed in mathematical statistics \cite{neutral,washington}|the
idea of the simplest density for {\it probability of probability}
allows to come up with the unique family of Dirichlet densities. This
family is characterized by a positive parameter $\beta$, which allows
quantitative comparison between phoneme frequencies for different
authors. From the physical side, the Dirichlet density is a direct
analogue of the ideal gas model from statistical mechanics, while
$\beta$ relates to the inverse temperature. Recall that the ideal-gas
model provides a simple and fruitful description of the
coarse-grained (thermodynamic) features of matter starting from the
principles of atomic and molecular physics \cite{balian}. Thus we
substantiate the atom-phoneme metaphor, that so far was developed only
qualitatively \cite{zwick,abler}.

Our second motivation for studying rank-frequency relations for
phonemes is whether they can provide information on the author of the
text, and thereby attempt at clarifying the psychological aspect of
phonemes. As seen below, the Dirichlet density not only leads to an
accurate description of phoneme rank-frequency relations, but it also
allows to establish that the frequencies of phonemes do depend on the
author of the text. We corroborate this result by an alternative
means.

The closest to the present approach is the study by Good \cite{good}
which was developed in
Refs. \cite{gusein_1,gusein_2,witten_bell}. These authors applied the
same idea on hidden probabilities as here, but they restricted
themselves by the flat density, which is a particular case $\beta=1$
of the Dirichlet density \cite{good,gusein_1,gusein_2}. Superficially,
this case seems to be special, because it incorporates the idea of
non-informative (unknown) probabilities (in the Bayesian sense)
\cite{jaynes}. However, the development of the Bayesian statistics has
shown that the $\beta=1$ case of the Dirichlet density is by no means
special with respect the prior information \cite{jaynes}. Rather, the
whole family of Dirichlet densities (with $\beta>0$ being a free
parameter) qualifies for this role \cite{schafer}.

This paper is organized as follows. Next section discusses the
Dirichlet density and its features. There we also deduce explicit
formulas for the probabilities ordered according to the Dirichlet
density. Then, we analyze the data obtained from English texts written
by different authors and show that it can be described via the
Dirichlet density. There we also demonstrate (in different ways
including non-parametric methods) that rank-frequency relations for
phonemes are author-dependent. Next, we show that the
author-dependency effect is not caused by common words used in
different texts. We summarize in the last section.

\section{Dirichlet density}
\label{diri}

\subsection{Ideal-gas models}

The general idea of applying ideal-gas models in physics \cite{balian}
is that for describing coarse-grained features of certain physical
systems, interactions between their constituents (atoms or molecules)
can be accounted for superficially (in particular, neglected to a
large extent). Instead, one focusses on the simplest statistical
description that contains only a few parameters (e.g. temperature,
volume, number of particles etc) \cite{balian}. In physics this
simplest descriptions amounts to the Gibbs density \cite{balian}. Its
analogue in mathematical statistics is known as the Dirichlet density,
and is explained below. Ideal gas models in physics are useful not
only for gases|where interactions are literally weak|but also for
solids, where interactions are important, but their detailed structure
is not, and hence it can be accounted for in a simplified way
\cite{balian}.

Following these lines, we apply below the Dirichlet density to phoneme
frequencies observed in a given text. More precisely, the ordered
frequencies generated by the Dirichlet density are compared with
observed (and ordered) frequencies of phonemes in a given text. This
ordering of frequencies amounts to a rough and simplified account of
(inter-phoneme) interactions, and suffices for an accurate description
of the rank-frequency relations for phonemes; see below.

\subsection{Definition and main features}

\begin{figure*}[htbp]
\begin{center}
    \includegraphics[width=0.5\columnwidth]{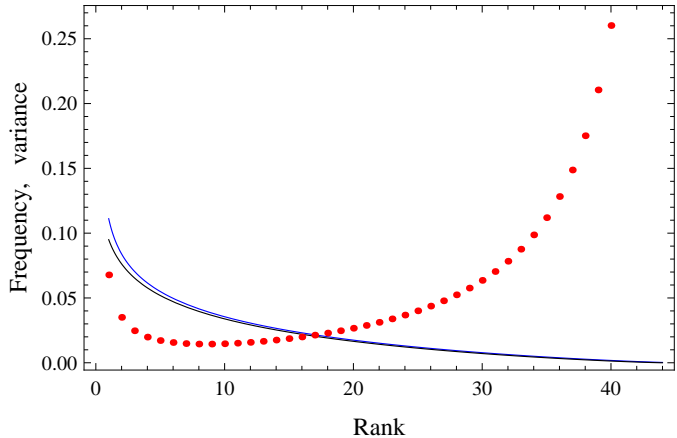}
    \caption{Rank-frequency curves and error generated by the
      Dirichlet density with $\beta=0.8$ and $n=44$. Blue curve:
      $\langle \theta_{(r)}\rangle$ (as a function of $r$) calculated
      according to (\ref{eq:a6}--\ref{eq:a5}). Black curve:
      $\hat{f}_r$ calculated via the approximate formula
      (\ref{eq:a14}); cf. section II of the supplementary
      material. Red points: the normalized variance $(\langle
      \theta_{(r)}^2\rangle -\langle \theta_{(r)}\rangle^2)/ \langle
      \theta_{(r)}\rangle^2$ for $r=1,...,44$ calculated according to
      (\ref{eq:a6}--\ref{eq:a5}). This expression is well approximated
      by (\ref{eq:a13}).}
\label{fig0}
\end{center}
\end{figure*}

The Dirichlet density ${\cal D}(\theta_1,...,\theta_n)$ is a
probability density over continuous variables
$(\theta_1,...,\theta_n)$ which by themselves have the meaning of
probabilities, i.e. ${\cal D}(\theta_1,...,\theta_n)$ is non-zero only
for $\theta_k\geq 0$ and $\sum_{k=1}^n\theta_k=1$:
\begin{eqnarray}
  \label{dirichlet}
{\cal D}(\theta_1,...,\theta_n|\beta_1,...,\beta_n)
=\frac{\Gamma[\sum_{k=1}^n
  \beta_k]}{\prod_{k=1}^n \Gamma[\beta_k]   }
\prod_{k=1}^n \theta_k^{\beta_k-1}\,
\delta(\sum_{k=1}^n \theta_k-1),
\end{eqnarray}
where $\beta_k>0$ are the parameters of the Dirichlet density,
$\delta(x)$ is the delta-function, $\Gamma[x]=\int_0^\infty \d \theta\, \theta^{x-1}e^{-\theta}$ is the Euler's $\Gamma$-function, and
(\ref{dirichlet}) is properly normalized: $\int_0^\infty \prod_{k=1}^n
\d \theta_k {\cal D}(\theta_1,...,\theta_n|\beta_1,...,\beta_n)=1$.

\comment{ We now recall features that determine the choice of
  (\ref{dirichlet}) as a density over probabilities; see
  \cite{washington} for a review. Each of these features indicates
  that the form of (\ref{dirichlet}) is stable with respect to certain
  transformations of probabilities. These transformations are natural
  for studying phonemes. An important feature that is not discussed
  here is that the Dirichlet density conserves its shape when updated
  via the multinomial conditional probability (and the Bayes formula)
  \cite{schafer}.}

The random variables $\Theta_1,...,\Theta_n$ (with realizations
$\theta_1,...,\theta_n$) are independent modulo the constraint that
they sum to $1$; see (\ref{dirichlet}). In this sense
(\ref{dirichlet}) is the simplest density for probabilities.  Now
(\ref{dirichlet}) for a particular case $\beta_k=\beta$ (which is most
relevant for our purposes) can be given the following
statistical-physics interpretation: if $\ln(\frac{1}{\theta_k})$ is
interpreted as the energy of $k$ \cite{shrejder,dover,vakarin}, then
$\beta-1$ becomes the inverse temperature for an ideal gas. It is
useful to keep this analogy in mind, when discussing further features
of the Dirichlet density.

Consider the subset $(\theta_1,...,\theta_m)$ ($m<n$) of probabilities
$(\theta_1,...,\theta_n)$. If $(\theta_1,...,\theta_m)$ should serve
as new probabilities, they should be properly normalized. Hence we
define new random variables as follows:
\begin{eqnarray}
  \label{eq:4}
(\widetilde{\theta}_1,...,\widetilde{\theta}_n)=
(\widehat{\theta}_1,...,\widehat{\theta}_m,\theta_{m+1},...\theta_{n}),
~~~
\widehat{\theta}_k=\frac{\theta_k}{\sum_{i=1}^m \theta_i}, ~~
k=1,...,m.
\end{eqnarray}
The joint probability ${\cal
  P}(\widetilde{\theta}_1,...,\widetilde{\theta}_n)$
now reads from
(\ref{dirichlet}):
\begin{eqnarray}
  \label{eq:5}
  {\cal
  P}(\widetilde{\theta}_1,...,\widetilde{\theta}_n)=
{\cal D}(\hat\theta_1,...,\hat\theta_m|\beta_1,...,\beta_m)\, {\cal X}
(\theta_{m+1},...,\theta_n),
\end{eqnarray}
where the precise form of ${\cal X}$ is not relevant for the message
of (\ref{eq:5}): if we disregard some probabilities and properly
re-normalize the remaining ones, the kept probabilities follow the
same Dirichlet density and are independent from the disregarded ones
\cite{washington}. This means that we do not need to know the number
of constituents before applying the Dirichlet density. This feature is
relevant for phonemes, because their exact number is to a large extent
a matter of convention, e.g. should English diphthongs be regarded as
separate phonemes, or as combinations of a vowel and a semi-vowel.

Condition (\ref{eq:5}) (called sometimes neutrality), together with
few smoothness conditions, determines the shape (\ref{dirichlet}) of
the Dirichlet density \cite{neutral}.

\comment{
{\bf 3.} This feature comes from aggregating probabilities, where two
(or more) outcomes are identified together and their probabilities are
summed. This is relevant for phonemes, since in the development of the
language distinct phonemes can develop into allophones of the same
phoneme. We have from (\ref{dirichlet}) for the density of the
remaining phonemes:
\begin{eqnarray}
  \label{eq:6}
 && \int_0^\infty\d\theta_{n-1}'\,\d
  \theta_{n}'\,\delta(\theta_{n-1}-\theta_{n-1}'-\theta_{n}')\,
  {\cal
    D}(\theta_1,...,\theta_{n-2},\theta_{n-1}',\theta_{n}'|\beta_1,...,\beta_n)
\nonumber\\
&&=  {\cal
    D}(\theta_1,...,\theta_{n-2},\theta_{n-1}|\beta_1,...,\beta_{n-2},
\beta_{n-1}+\beta_n).
\end{eqnarray}
This feature shows that $\beta_k>0$ can be regarded as a
(unnormalized) weight of the probability $\theta_k$. Generalizing to
aggregation of more than two probabilities is obvious.
}

Assuming $n$ free parameters $\beta_k$ for $n$ phoneme frequencies does
not amount to any effective description. Hence below we employ
(\ref{dirichlet}) with
\begin{eqnarray}
  \label{eq:3}
  \beta_k=\beta,
\end{eqnarray}
for describing the ranked phoneme frequencies.  This implies that the
full vector $(\beta_1,...,\beta_{n})$ is replaced by a certain
characteristic value $\beta$, which is to be determined from comparing
with data. To provide some intuition on $\beta$, let us note from
(\ref{dirichlet}) that a larger value of $\beta$ leads to
more homogeneous density (many events have approximately equal
probabilities). For $\beta_k\to 0$ the region $\theta_k\simeq 0$ is the
most probable one.

\comment{
{\bf 4.} Consider the multi-nomial density with probabilities
$\theta_1,...,\theta_n$:
\begin{eqnarray}
  \label{eq:8}
  {\cal
    P}(\nu_1,...,\nu_n|
  \theta_1,...,\theta_n)=\frac{(\sum_{k=1}^n\nu_k)!}
{\prod_{k=1}^n\nu_k!}\, \theta_1^{\nu_1}...\theta_n^{\nu_n},
\end{eqnarray}
which is the probability of seeing (in $\sum_{k=1}^n\nu_k$ independent
trials) $\nu_k$ times the event $k$.
If $(  \theta_1,...,\theta_n)$ are unknown, and we are given a new
data $\{\nu_k\}_{k=1}^n$, then the prior Dirichlet density
conserves its shape under updating via the Bayes formula
\cite{jaynes}:
\begin{eqnarray}
  \label{eq:9}
 {\cal
    P}(\theta_1,...,\theta_n|\nu_1,...,\nu_n)&=&
  \frac{  {\cal
      P}(\nu_1,...,\nu_n|\theta_1,...,\theta_n)
{\cal D}(\theta_1,...,\theta_n|\beta_1,...,\beta_n)}{
\int\prod_{k=1}^n\d \theta'_k{\cal
      P}(\nu_1,...,\nu_n|\theta'_1,...,\theta'_n)
{\cal D}(\theta'_1,...,\theta'_n|\beta_1,...,\beta_n)
} \\
  \label{eq:10}
&=&
{\cal
    D}(\theta_1,...,\theta_n|\beta_1+\nu_1,...,\beta_n+\nu_n).
\end{eqnarray}
}

\subsection{Distribution of ordered probabilities
(order statistics)}

The random variables $\Theta_1,...,\Theta_n$ (whose realizations are
$\theta_1,...,\theta_n$ in (\ref{dirichlet})) are now put in a
non-increasing order:
\begin{eqnarray}
  \label{eq:7}
  \Theta_{(1)}\geq...\geq\Theta_{(n)}.
\end{eqnarray}
This procedure defines new random variables, so called order
statistics of the original ones \cite{david}. We are interested by the
marginal probability density of $\Theta_{(r)}$. It is difficult to
obtain this object explicitly, because the initial
$\Theta_1,...,\Theta_n$ are correlated random variables. However, we
can explicitly obtain from (\ref{dirichlet}) a two-argument function
that suffices for calculating the moments of $\Theta_{(r)}$
[see section II of the supplementary material]:
\begin{eqnarray}
  \label{eq:a6}
\chi_r(y;m)=\frac{\Gamma[n\beta]}{\Gamma[n\beta+m]}\,
\frac{n!}{(n-r)!(r-1)!}\, \frac{y^{\beta-1}e^{-y}}{\Gamma[\beta]}\,
\varphi^{n-r}(y)
[1-\varphi(y)]^{r-1} ,
\end{eqnarray}
where $\Gamma[x]$ is the $\Gamma$-function and where
\begin{eqnarray}
  \label{eq:a7}
  \varphi (y)=\frac{1}{\Gamma[\beta]}\int_0^y\d x\, x^{\beta-1}e^{-x}
\end{eqnarray}
is the regularized incomplete $\Gamma$-function. Now the moments of
$\Theta_{(r)}$ are obtained as
\begin{eqnarray}
  \label{eq:a5}
\langle \theta_{(r)}^m\rangle =\int_0^\infty \d y\, y^m   \chi_r(y;m).
\end{eqnarray}

\begin{figure*}[htbp]
\centering
    \includegraphics[width=0.95\columnwidth]{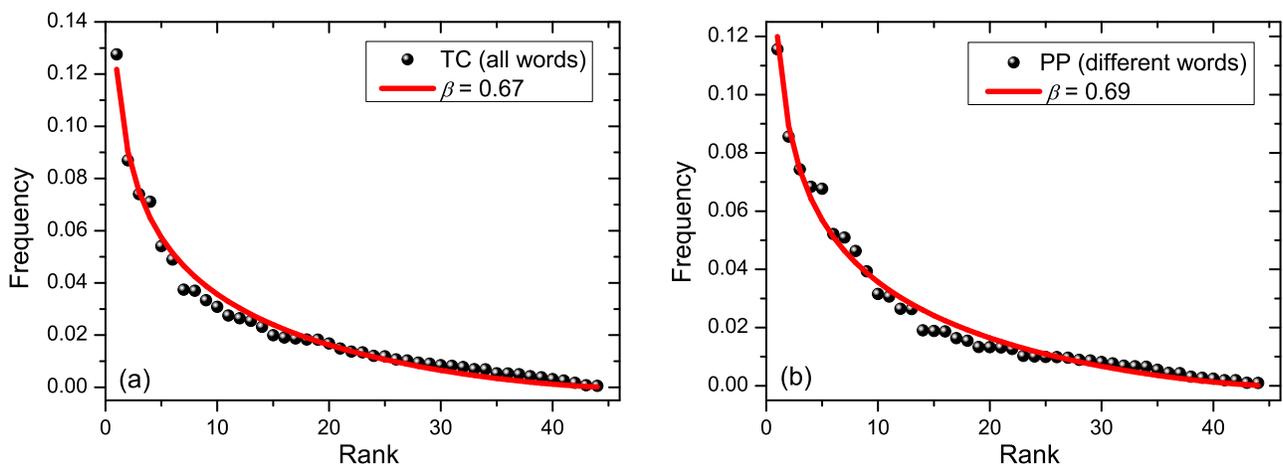}
    \caption{Rank-frequency relation (black circles) and the fitting
      with Dirichlet distribution (red line). (a) Left figure: text TC,
      where frequencies were extracted from all words. (b) Right
      figure: text PP, where different words were employed; see Table
      I for the description of texts.}
\label{fig1}
\end{figure*}

\begin{figure*}[htbp]
\centering
   \includegraphics[width=0.95\columnwidth]{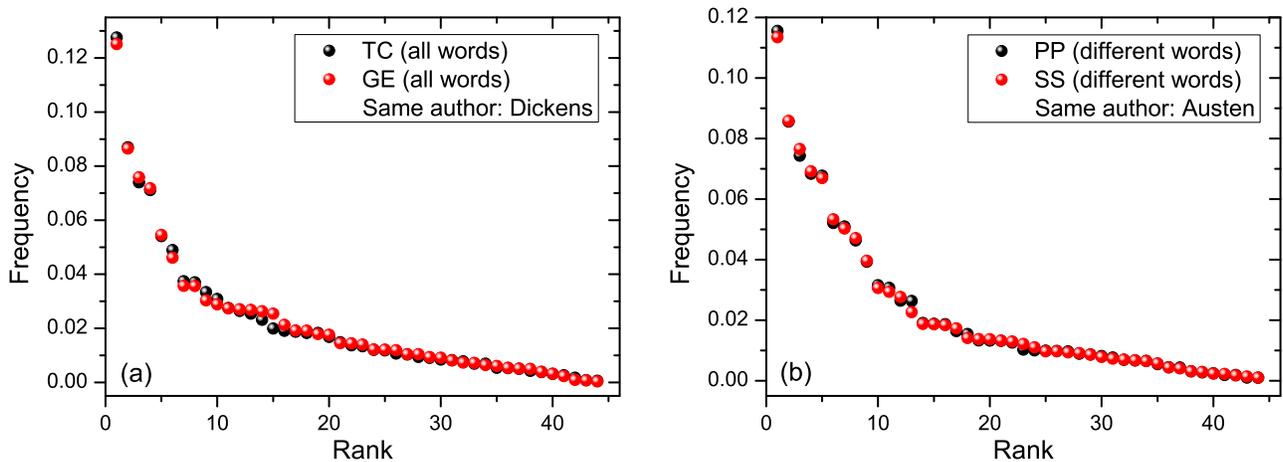}
    \caption{Rank-frequency relation (black and red circles) for two
      texts written by the same author. (a) Left figure: TC and GE written by
      Dickens (all words were employed for extracting the phoneme
      frequencies). (b) Right figure: PP and SS written by Austen (different words
      were employed); see Table I.
    }
\label{fig2}
\end{figure*}

\begin{figure*}[htbp]
\centering
 \includegraphics[width=0.95\columnwidth]{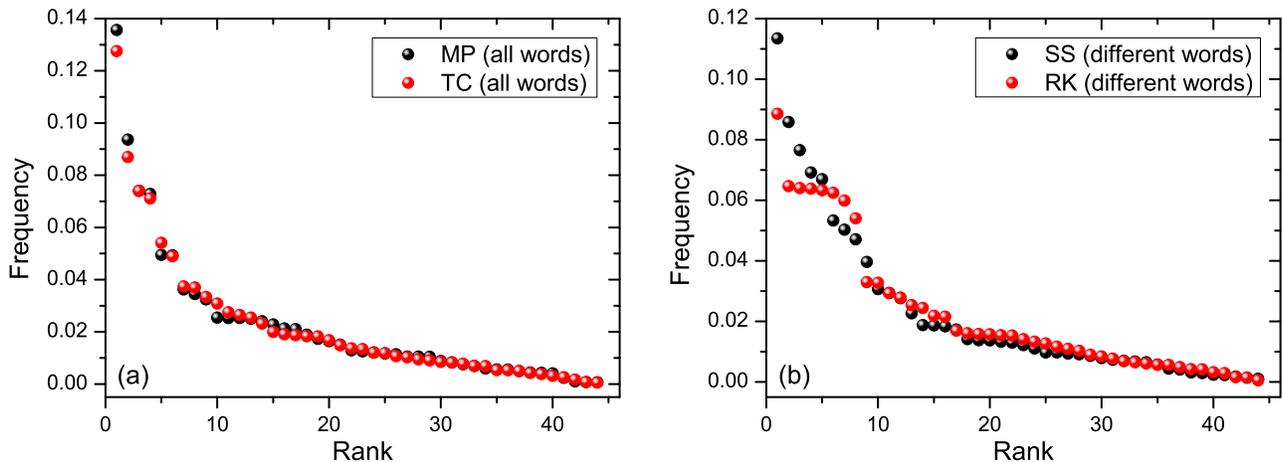}
 \caption{ Rank-frequency relation (black and red circles) for two
   texts written by different authors. (a) Left figure: TC by Dickens {\it
     versus} MP by Austen (all words were employed). (b) Right figure: SS
   by Austen {\it versus} RK by Tolkien (different words were
   employed); see Table I for parameters of these texts.  }
\label{fig3}
\end{figure*}

In the next section we shall see that the sequence of ordered
probabilities $f_r$ [cf. (\ref{eq:0})] can be generated via
(\ref{eq:a6}). To this end, the empiric quantities $f_r$ will be
compared to $\hat{f}_r=\langle \theta_{(r)}\rangle$;
cf. (\ref{eq:a5}). The rationale for using the average is that for
parameters we are interested in|where $n\simeq 40-50$ (for English
phonemes $n=44$) and $0.5\leq \beta\leq 1$|we get from
(\ref{eq:a6}--\ref{eq:a5}) that relative fluctuations around the
average $\hat{f}_r\equiv\langle \theta_{(r)}\rangle$ are
small. Namely, $\varepsilon_r \equiv(\langle \theta_{(r)}^2\rangle
-\langle \theta_{(r)}\rangle^2)/ \langle \theta_{(r)}\rangle^2\lesssim
0.02$ for all values of $r$, excluding $r\approx n$, i.e. very low
frequency phonemes. This is shown in Fig.~\ref{fig0} for a particular
value $\beta=0.8$. Note that $\varepsilon_r$ is not a monotonic
function of $r$: it is smallest for middle ranks. (Even for those
values of $r$, where $\varepsilon_r\simeq 1$, $\hat{f}_r=\langle
\theta_{(r)}\rangle$ can still describe the empiric frequencies $f_r$,
as seen below.) Now there is a simpler approximate formula for
$\hat{f}_r=\langle \theta_{(r)}\rangle$ that is deduced from
(\ref{eq:a5}) [see section II of the supplementary material]:
\begin{eqnarray}
  \label{eq:a14}
  \frac{r}{n}=1-\varphi(\hat{f}_r n\beta).
\end{eqnarray}

Fig.~\ref{fig0} shows that $\hat{f}_r$ obtained from (\ref{eq:a14})
indeed approximates well $\langle \theta_{(r)}\rangle$ for almost all
ranks $r$.

\section{Results}

\label{resu}

\subsection{Fitting rank-frequency relations to the Dirichlet
  distribution}

\begin{table*}[htbp]
\begin{center}
  \caption{ Nine texts and their parameters. Texts are abbreviated and
    numbered. $N_{tw}$, $N_{pht}$, $N_{dw}$ and $N_{phd}$ are, respectively, the
    total number of words, the number of phonemes of the total words, the number of different words and the
    number of phonemes of different words. \\
    J. Austen: {\it Mansfield Park} (MP or 1) 1814; {\it Pride and
      Prejudice} (PP or 2) 1813; {\it Sense and Sensibility} (SS or 3) 1811. \\
    C. Dickens: {\it A Tail of Two Cities} (TC or 4) 1859; {\it Great
      Expectations} (GE or 5) 1861; {\it Adventures of Oliver
      Twist} (OT or 6) 1838.\\
    J. Tolkien: {\it The Fellowship of the Ring} (FR or 7) 1954; {\it
      The Return of the King} (RK or 8) 1955; {\it The
      Two Towers} (TT or 9) 1954.
  }
\begin{tabular}{||c|c|c|c|c||}
\hline\hline
Texts & $N_{tw}$ & $N_{pht}$ &   $N_{dw}$ & $N_{phd} $    \\
\hline\hline
MP (1) & 160473 & 567750 &  7854 &  48747    \\
PP (2) & 121763 & 435322 & 6385 &  39767    \\
SS (3) & 119394 & 425822 & 6264 &  38668    \\
\hline
TC (4) & 135420 & 468642 &9841   &  58760 \\
GE  (5) & 186683 & 623079 & 10933  &  65364  \\
OT  (6)& 159103 & 555372 &10359  &  61072   \\
\hline
FR (7) & 177227 & 617106 &8644 &  46509    \\
TT (8) & 143436 & 502303 &7676 &  39823    \\
RK (9) & 134462 & 431141 &7087 &  36494    \\
\hline\hline
\end{tabular}
\end{center}
\end{table*}

\begin{table*}[htbp]
\begin{center}
\tabcolsep0.060in \arrayrulewidth0.3pt
\caption{Fitting parameters for texts numbered as 1-9; see
  (\ref{eq:1}, \ref{eq:2}) and Table I for text numbers. The phoneme
  frequencies are extracted from all words of the text; see Table III
  for the values of $\beta$ calculated from different words of texts.}
\begin{tabular}{||c||c|c|c||c|c|c||c|c|c||}
\hline\hline
   & 1  & 2  & 3   & 4 & 5 & 6 & 7 & 8 & 9    \\
\hline\hline
$\beta$ &0.61  &0.63   &0.61    &0.67  &0.69 &0.69  &0.75   &0.74   &0.79    \\
\hline
$SS_{err}\times 10^7$ &7696  &7574   &6151    &4317  &5287 &3993   &4196   &4337 &3580     \\
\hline
$R^2$ &0.9768  &0.9765   &0.9816    &0.9859  &0.9820 &0.9867  &0.9844   &0.9842 &0.9860     \\
\hline\hline
\end{tabular}
\end{center}
\end{table*}

\begin{table*}[htbp]
\begin{center}
\tabcolsep0.060in \arrayrulewidth0.3pt
\caption{Fitting parameters for texts numbered as 1-9; see
  (\ref{eq:1}, \ref{eq:2}) and Table I for text numbers. The phoneme
  frequencies are extracted from different words of the text; see
  Table II for the values of $\beta$ calculated from all words of
  texts. Eqs.~(\ref{eq:18}, \ref{eq:15}) compare the data presented in
  Tables II and III.}
\begin{tabular}{||c||c|c|c||c|c|c||c|c|c||}
\hline\hline
   & 1  & 2  & 3   & 4 & 5 & 6 & 7 & 8 & 9    \\
\hline\hline
$\beta$ &0.72  &0.69   &0.69    &0.77  &0.78 &0.79 &0.968 &0.979  &0.975  \\
\hline
$SS_{err}\times 10^7$ &5150  &4495   &5003  &6107  &5265   &5220
&11296   &12943  &10366  \\
\hline
$R^2$ &0.9818  &0.9847   &0.9829    &0.9771  &0.9800   &0.9800
&0.9501 &0.9403    &0.9525  \\
\hline\hline
\end{tabular}
\end{center}
\end{table*}

We studied 48 English texts written by 16 different, native-English
authors; see Table I and section III of the supplementary
material. For each text we extracted the phoneme frequencies
$\{f_r\}_{r=1}^n$ and ordered them as in (\ref{eq:0}); the list of
English phonemes is given in section I of the supplementary material.
The transcription of words into phonemes was carried out via the software
PhoTransEdit, which is available at \cite{pho}. This is a relatively
slow, but very robust software, since it works by checking each word
in the phonetic dictionary. Thus it can err only on those unlikely cases, when
the word is not found in the dictionary.

It is important to specify from which set of words (of a text) one
extracts the phoneme frequencies. Two natural choices are possible
here: either one employs all words of the text, or different words of
the text (i.e. multiple occurrences of the same word are
neglected). We shall study both cases. For clarity reasons, we shall
present our results by focussing on the three authors mentioned in
Table I. Three texts by three authors is in a sense the minimal set-up
for described effects.  We stress that other texts we studied fully
corroborate our results; they are partially described in Table
\ref{tabu} below and in section III of the supplementary material.

The ordered set $\{f_r\}_{r=1}^n$ of phoneme frequencies for each text
was compared with the prediction $\{\hat{f}_r=\langle
\theta_{(r)}\rangle \}_{r=1}^n$ of the Dirichlet density [see
(\ref{eq:a5})]. Here the parameter $\beta$ [cf. (\ref{dirichlet},
\ref{eq:3})] is found from minimizing the error:
\begin{eqnarray}
  \label{eq:1}
  SS_{err}=\sum_{k=1}^n (f_k-\hat{f}_k)^2.
\end{eqnarray}
For each studied case we also monitored the coefficient of correlation
between $\{f_r\}_{r=1}^{n}$ and $\{\hat{f}_r\}_{r=1}^{n}$:
\begin{eqnarray}
  \label{eq:2}
R^2=\frac{  \left[\, \sum_{k=1}^n (f_k-\bar{f})
(\hat{f}_k-\overline{\hat{f}})\, \right]^2    }{\sum_{k=1}^n
(f_k-\bar{f})^2 \sum_{k=1}^n
(\hat{f}_k-\overline{\hat{f}})^2},
\end{eqnarray}
where
\begin{eqnarray}
  \label{eq:11}
\bar{f}\equiv \frac{1}{n}{\sum}_{k=1}^n f_k, \qquad
\overline{\hat{f}}\equiv \frac{1}{n}{\sum}_{k=1}^n \hat{f}_k.
\end{eqnarray}
A good fitting means that $R^2$ is close to $1$. We found that (as
functions of $\beta$) $ SS_{err}$ and $1-R^2$ minimize simultaneously.

Examples of fitting curves for phoneme rank-frequency relations are
presented in Fig.~\ref{fig1}. The fitting parameters are given in
Tables II and III. Note that the fitting values of $R^2$ are good.
The group of most frequent eight phonemes reads [see section I of
the supplementary material]: /\i/,\, /\textschwa/,\, /n/,\, /s/,\,
/t/,\, /l/,\, /d/,\, /r/. The concrete ranking between them depends on
the text, but the most frequent one is normally /\i/.

Tables II and III show that the texts by the same author have closer
values of $\beta$ than those written by different authors; see also
Figs.~\ref{fig2} and \ref{fig3}. This can be quantified via the
following three inequalities
\begin{eqnarray}
  \label{eq:151}
&&  0 < b({\rm A})\equiv
  {\rm min\,}\{ |\beta_i-\beta_k| \}_{i=1,2,3; k=4,5,6,7,8,9}\, -\, {\rm
    max\,}\{|\beta_i-\beta_j|\}_{i<j; i,j=1,2,3},   \\
  \label{eq:152}
&& 0< b({\rm D})\equiv
  {\rm min\,}\{   |\beta_i-\beta_k| \}_{i=4,5,6; k=1,2,3,7,8,9}\,-\, {\rm
    max\,}\{|\beta_i-\beta_j|\}_{i<j; i,j=4,5,6},    \\
  \label{eq:153}
&&0< b({\rm T})\equiv
  {\rm min\,}\{|\beta_i-\beta_k| \}_{i=7,8,9; k=1,2,3,4,5,6} \,-\,  {\rm
    max\,}\{|\beta_i-\beta_j|\}_{i<j; i,j=7,8,9},
\end{eqnarray}
where A, D and T refer, respectively to Austen, Dickens and Tolkien
[see Table I]. The indices $i$ and $j$ run over the texts by the same
author, while $k$ refer to different authors, e.g.  $i,j=\{1,2,3\}$
(Austen) and $k=\{4,5,6,7,8,9\}$ (not Austen). The minimization (or
maximization) in (\ref{eq:151}-\ref{eq:153}) goes over indicated
indices.

Eqs.~(\ref{eq:151}-\ref{eq:153}) hold both for phoneme frequencies
extracted from different words and from all words of a text;
cf. Tables II and III. For instance, $b^{\rm [all~words]} ({\rm
  A})=0.02$, $b^{\rm [all~words]} ({\rm D})=0.02$, $b^{\rm
  [all~words]} ({\rm T})=0$. The latter is the only minor exclusion
from (\ref{eq:151}-\ref{eq:153}).

Thus the set $\{\beta_i\}_{i=1}^9$ fragments into
three clusters that refer to different authors. Note that
\begin{eqnarray}
  \label{eq:18}
  b^{\rm [diff.~words]} (a) > b^{\rm [all~words]} (a),
  \qquad   a={\rm A, \, D, \, T}.
\end{eqnarray}
Hence different words display the author-dependency in a stronger
form; this is confirmed below by other methods.

The author-dependency of phoneme rank-frequency relation is
unexpected, because the rank-frequency relation for words (which
consists of phonemes) follows the Zipf's law whose shape is
independent of the author \cite{wyllis,baa,zipf_pre}.
Note that the few most frequent phonemes and the least frequent ones
appear to fit best the theoretical prediction; cf. Fig.~\ref{fig1}.
This feature again contrasts the rank-frequency relation for
words, where it is known that high-frequency words|these are mostly
the functional words, e.g. {\it and}, {\it or}|do hold the Zipf's law
worst than other words \cite{zipf_pre}. On the other hand, the
moderate-frequency phonemes deviate most from the prediction of the
Dirichlet curve; cf. Fig.~\ref{fig1}. This effect is not statistical,
since fluctuations around the average are most suppressed for
moderate-frequency phonemes; see after (\ref{eq:a5})
and Figs.~\ref{fig0} and \ref{fig1}.

Another pertinent result is that [see Tables II and III]
\begin{eqnarray}
  \label{eq:15}
  \beta_i^{\rm [diff.~words]} > \beta_i^{\rm [all~words]},~\qquad i=1,...,9,
\end{eqnarray}
i.e. the phoneme distribution obtained from different words is more
homogeneous [see our discussion after (\ref{eq:3})], because for all
words the frequency of high-rank phonemes is amplified due to multiple
usage of frequent words.

Note that the above three texts belong to one genre (novels) and
concern only three authors. Hence we studied other 13 native-English
authors who created in XIX'th and the first half of
XX'th century; see section III of the supplementary material. These
additional studies corroborate the obtained results. In particular,
Table~\ref{tabu} presents the values of $\beta$ extracted from
different texts of 5 authors. These authors were selected so that
their language differences due to social, temporal and professional
backgrounds are minimized.  In addition, we selected 4 of them to be
professional scientists, since the language of scientific works is
normally more unified.  Lyell, Darwin, Wallace, and Spenser were
naturalists, while the fifth author (H.G. Wells) held a PhD in
biology and wrote a lot about scientists.  Lyell strongly influenced
Darwin, while Darwin and Wallace were close colleagues. All these
three naturalists influenced Spenser and Wells. However,
Table~\ref{tabu} shows that the values of $\beta$ for these 5 authors
are clearly different and hold analogues of
(\ref{eq:151}-\ref{eq:153}).

We want to stress that $\beta$ anyhow changes in a bounded interval:
$0.5<\beta<1$. Hence if one takes sufficiently many authors, their
values of $\beta$ will start to overlap. In our study of (overall) 16
authors we confirmed this expectation; see section III of the
supplementary material. However, these overlaps are accidental,
i.e. the overlapping authors can be easily distinguished by
alternative means. In particular, their phoneme distributions can be
robustly distinguished via distances, as described below.

\begin{table*}[htbp]
\begin{center}
\tabcolsep0.060in \arrayrulewidth0.3pt
\caption{The values of $\beta$ extracted from different words of texts
  for 5 authors. For each author we analyzed three texts. They are
  described in the supplementary material, where we also discuss 8
  other authors.  }
\begin{tabular}{||c||c|c|c||}
\hline\hline
C. Lyell
& 0.798  & 0.785  & 0.792  \\
\hline\hline
A. R. Wallace
& 0.744  & 0.756  & 0.739  \\
\hline\hline
C. Darwin
& 0.817  & 0.810  & 0.822  \\
\hline\hline
H. Spenser
& 0.646   & 0.658  & 0.650  \\
\hline\hline
H. G. Wells
& 0.737  & 0.735  & 0.724  \\
\hline\hline
\end{tabular}
\label{tabu}
\end{center}
\end{table*}

\subsection{Distance between phoneme frequencies}

The author-dependency of phoneme rank-frequency relation is
corroborated by looking directly at suitable distances between the
ranked phoneme frequencies in different texts. We choose to work with
the variational distance
\begin{eqnarray}
  \label{4}
  \rho_1(ij)=\frac{1}{2}\sum_{k=1}^n
  |\,f_k[i]-f_k[j]\,|,
\end{eqnarray}
where $\{f_k[i]\}_{k=1}^n$ are the ordered phoneme frequencies in the
text $i$. We shall also employ a more fine-grained (detail-specific)
distance. Let $f[\alpha|i]$ be the frequency of phoneme $\alpha$ in
text $i$ ($\alpha=1,...,n$, $i=1,..,9$). We can now define [cf.~(\ref{4})]
\begin{eqnarray}
  \label{1}
  \rho_0(ij)=\frac{1}{2}\sum_{\alpha=1}^n
  |\,f[\alpha|i]-f[\alpha|j]\,|.
\end{eqnarray}
Now $\rho_0(ij)=0$ only if $f[\alpha|i]=f[\alpha|j]$. It is seen from
Tables V-VII that $\rho_0(ij)>\rho_1(ij)$, as it
should be, because $\rho_1(ij)$ is less sensitive to details (i.e. it
is more coarse-grained).

To motivate the choice of the variational distance
$\rho_0=\frac{1}{2}\sum_{\alpha=1}^n |\,p_\alpha -q_\alpha\,|$ between
two sets of probabilities $\{p_\alpha\}_{\alpha=1}^n$ and
$\{q_\alpha\}_{\alpha=1}^n$, let us recall an important feature of
this distance \cite{gibbs}: $\rho_0 ={\rm
  max}_{\Omega}\left|\sum_{\alpha\in \Omega}
  (p_\alpha-q_\alpha)\,\right|$, where the maximization goes over all
sub-sets $\Omega$ of $\{1,...,n\}$. Thus $\rho_0$ refers to the
(composite) event that gives the largest probability difference
between $\{p_\alpha\}_{\alpha=1}^n$ and $\{q_\alpha\}_{\alpha=1}^n$.

Tables V and VI refer, respectively, to phoneme frequencies extracted
from all words and different words of the text. These tables show that
phoneme rank-frequency relations between the texts written by the same
author are closer to each other|in the sense of distances $\rho_0$ and
$\rho_1$|than the ones written by different authors. This is also seen
on Figs.~\ref{fig2} and \ref{fig3}.

To quantify these differences, consider the following inequalities
that define clustering with respect to authors (see Table I for
numbering of texts, and note that $\rho_\lambda(ij)=\rho_\lambda(ji)$
for the distance between the texts $i$ and $j$):
\begin{eqnarray}
  \label{eq:12}
&&  0 < z_\lambda({\rm A})\equiv
  {\rm min\,}\{\rho_\lambda(ik)\}_{i=1,2,3; k=4,5,6,7,8,9}\, -\, {\rm
    max\,}\{\rho_\lambda(ij)\}_{i<j; i,j=1,2,3},  ~~ \lambda=0,1, \\
  \label{eq:13}
&& 0< z_\lambda({\rm D})\equiv
  {\rm min\,}\{\rho_\lambda(ik)\}_{i=4,5,6; k=1,2,3,7,8,9}\,-\, {\rm
    max\,}\{\rho_\lambda(ij)\}_{i<j; i,j=4,5,6},   ~~ \lambda=0,1, \\
  \label{eq:14}
&&0< z_\lambda({\rm T})\equiv
  {\rm min\,}\{\rho_\lambda(ik)\}_{i=7,8,9; k=1,2,3,4,5,6} \,-\,  {\rm
    max\,}\{\rho_\lambda(ij)\}_{i<j; i,j=7,8,9},   ~~
  \lambda=0,1,
\end{eqnarray}
where A, D and T refer, respectively to Austen, Dickens and Tolkien;
cf.~(\ref{eq:12}--\ref{eq:14}) with (\ref{eq:151}-\ref{eq:153}). For
example, the maximal distance (\ref{1}) between texts by Austen (see
Table I) is denoted by ${\rm max\,}\{\rho_0(ij)\}_{i<j; i,j=1,2,3}$,
while ${\rm min\,}\{\rho_0(kl)\}_{k=1,2,3; l=4,5,6,7,8,9}$ is the
minimal distance between texts written by Austen and those written by
Dickens and Tolkien. Note that (\ref{eq:12}--\ref{eq:14}) hold as well
for other 13 authors we analyzed; see section III of the supplementary
material for examples.

The meaning of (\ref{eq:12}--\ref{eq:14}) can be clarified by looking
at an authorship attribution task: let several texts $i=1,2,3$ by
(for example) Austen are at hands, and one is given an unknown text
$\alpha$. The question is whether $\alpha$ could also be written by
Austen. If now ${\rm max}_i[\rho_\lambda(i\alpha)] \leq {\rm
  max}_{i<j}[\rho_\lambda(ij)]$, we have an evidence that $\alpha$ is
written by Austen.

We stress that there are no fitting parameters in
(\ref{1}--\ref{eq:14}). Our data (cf. Tables V and VI) holds eleven
(out of twelve) inequalities (\ref{eq:12}--\ref{eq:14}) for phoneme
frequencies extracted both from different and from all words of the
text. There is only one exclusion: $z_1^{\rm [all~words]} ({\rm
  T})=-0.00207$, which is by an order of magnitude smaller than the
respective frequencies [cf.~(\ref{eq:14})]. Apart of this minor
exclusion, we confirm the above prediction (obtained via the fitted
values of $\beta$) on the author-dependency for phoneme frequencies.

Data shown in Tables V (all words) and VI (different words) also imply
the following inequalities [confirming (\ref{eq:18})]
\begin{eqnarray}
  \label{eq:20}
  z_\lambda^{\rm [diff.~words]} (a) > z_\lambda^{\rm [all~words]} (a),
  \qquad \lambda=0,1, \qquad
  a={\rm A, \, D, \, T}.
\end{eqnarray}

Another pertinent feature is that the distances $\rho_0$ and $\rho_1$
between texts written by the same author hold
\begin{eqnarray}
  \label{eq:16}
&& \rho^{\rm [all~words]}_\lambda(ij) > \rho^{\rm
[diff.~words]}_\lambda(ij),\\
&& \lambda=0,1, \quad
(ij)=\{\, (12), (13), (23), (45), (46), (78), (79), (89)\,  \}.  \nonumber
\end{eqnarray}
Seventeen out of eighteen relations (\ref{eq:16}) hold for our data;
see Tables V and VI, where we present the distances $\rho_0$ and
$\rho_1$ for phoneme frequencies deduced from, respectively, all words
and different words of the texts. The only exclusion in (\ref{eq:16})
is $\rho^{\rm [diff.~words]}_0(78)-\rho^{\rm
  [all~words]}_0(78)=0.02853-0.02584=0.00269$. No definite relations
exist between $\rho^{\rm [all~words]}_\lambda$ and $\rho^{\rm
  [diff.~words]}_\lambda$ for texts written by different authors.  One
can interpret (\ref{eq:16}) as follows. When going from different
words to all words of the text, the majority of frequent words are not
author-specific: they are mostly key-words (that are specific to the
text, but not necessarily to the author) and functional words
(e.g. {\it and}, {\it or}, {\it of}, {\it but}) that are again not
author-specific.

Taken together, (\ref{eq:20}) and (\ref{eq:16}) imply that the clustering
with respect to authors is better visible for frequencies extracted
from different words of the texts (the inter-cluster distance increases,
whereas the intra-cluster distance decreases). The same effect was
obtained above via fitted values of $\beta$'s; see (\ref{eq:18}).

\comment{
The frequencies are ordered as in (\ref{eq:0}).
The difference between $S_1$ and $S_2$ can be studied from the
viewpoint of the ordering induced by (\ref{eq:0}):
\begin{eqnarray}
  \label{3}
  \rho_1(S_1,S_2)=\sum_{\alpha=1}^n
  |\,R[\alpha|S_1]-R[\alpha|S_2]\,|,
\end{eqnarray}
where $R[\alpha|S_i]$ is the rank (defined via (\ref{eq:0})) of
phoneme $\alpha$ in $S_i$. Now $ \rho_1(S_1,S_2)$ nullifies
\footnote{The maximal value of $\rho_1(S_1,S_2)$ is reached when the
  two ordering are completely opposite. This maximal value is equal to
  $\frac{n^2-1}{2}$, if $n$ is odd, and to $\frac{n^2}{2}$ if $n$ is
  even. } only if the notion of more frequent (phoneme) is the same
for both $S_1$ and $S_2$; it is not needed that they have exactly the
same frequencies both on $S_1$ and $S_2$. }

\begin{table*}[htbp]
\begin{center}
\tabcolsep0.030in \arrayrulewidth0.3pt
\caption{Distances $\rho_0$ and $\rho_1$ between texts; see Table I
  and (\ref{1}--\ref{4}) for the definition of $\rho_0$ and
  $\rho_1$. The phoneme frequencies are extracted from all words of
  the text. Eqs.~(\ref{eq:20}, \ref{eq:16}) compare the distances from
all words with those from different words. }
\begin{tabular}{||c||c|c|c||c|c|c||c|c|c||}
\hline\hline
   & 12  & 13  & 23   & 45 & 46 & 56 & 78 & 79 & 89    \\
\hline\hline $\rho_0\times 10^5$ & 3045 &  2062 &  2549  & 3423
& 2382  &
3448  & 2584  & 2066 & 2464 \\
\hline $\rho_1\times 10^5$ & 2227 & 1602 & 2103  &2100 &1978  &
2753  & 1808  & 1809 & 2037 \\ \hline \hline
\end{tabular}
\begin{tabular}{||c|c|c|c|c|c|c|c|c|c|c|c|c||}
\hline\hline
   & 14 & 15 & 16 & 17 & 18 & 19 & 24 & 25 & 26 & 27 & 28 & 29    \\
\hline\hline $\rho_0\times 10^5$ & 3583 &4690 &4000  &7372
&7402 &
7322 &3645 &4762 &4064 &7653 &7629  &7650 \\
\hline $\rho_1\times 10^5 $ & 2784 &3044 &3260  &5149 &5227  &
5599  &2712  &3059  &3110 &4978 &5052  &5449 \\
\hline\hline
\end{tabular}
\begin{tabular}{||c|c|c|c|c|c|c|c|c|c|c|c|c|c|c|c||}
\hline\hline
   & 34 & 35 & 36 & 37 & 38 & 39 & 47 & 48 & 49 & 57 & 58 & 59 & 67&
   68& 69    \\
\hline\hline $\rho_0\times 10^5$ & 3562 &  4924 &  4358  & 7737
&6950 &7447 &5174 &5327 &5061 &6113 &6436
&6217  &5074  &5706  &5202\\
\hline $\rho_1\times 10^5$ &2546  &3022  &3181  &5266  &5085
&5654 &3950
&3568  &3935   &3894 &4014  &4325  &3727  &3934  &3770 \\
\hline\hline
\end{tabular}
\end{center}
\end{table*}


\begin{table*}[htbp]
\begin{center}
\tabcolsep0.030in \arrayrulewidth0.3pt
  \caption{Distances $\rho_0$ and $\rho_1$ between texts; see Table I
    and (\ref{1}--\ref{4}). The phoneme frequencies are extracted from
    different words of the text; see (\ref{eq:20}, \ref{eq:16}) for
    comparison with all words.}
\begin{tabular}{||c||c|c|c||c|c|c||c|c|c||}
\hline\hline
   & 12  & 13  & 23   & 45 & 46 & 56 & 78 & 79 & 89    \\
\hline\hline $\rho_0\times 10^5$ &1563  &1317  &1413  &1568
&1380 &1100  &2853  &1946  &2025
\\
\hline $\rho_1\times 10^5$ &1346  &1205  &1346 &1266 &1126
&1052 &1635  &1476  &1569
\\
\hline\hline
\end{tabular}
\begin{tabular}{||c|c|c|c|c|c|c|c|c|c|c|c|c||}
\hline\hline
   & 14 & 15 & 16 & 17 & 18 & 19 & 24 & 25 & 26 & 27 & 28 & 29    \\
\hline\hline
$\rho_0\times 10^5$ 
&2296 &2703 &2868 &7430 &9535 &8434 &2839 &3318 &3458 &8141 &9999
&9167
\\
\hline
$\rho_1\times 10^5$ 
&1967 &2110 &2470 &6103 &7200 &6775 &2252 &2436 &2709 &6587 &7544
&7136
\\
\hline\hline
\end{tabular}
\begin{tabular}{||c|c|c|c|c|c|c|c|c|c|c|c|c|c|c|c||}
\hline\hline
   & 34 & 35 & 36 & 37 & 38 & 39 & 47 & 48 & 49 & 57 & 58 & 59 & 67&
   68& 69    \\
\hline\hline
$\rho_0\times 10^5$ 
&2718 &3264  &3257  &7943 &9998  &8997  &5918 &7875  &6899  &5521
&7842  &6646  &5595  &7785  &6786
\\
\hline
$\rho_1\times 10^5$ 
&2193  &2486 &2636 &6539 &7447 &7022 &4795 &5971 &5368 &4631
&5566 &5222 &4486 &5645  &5201
\\
\hline\hline
\end{tabular}
\end{center}
\end{table*}

\begin{table*}[htbp]
\begin{center}
\tabcolsep0.010in \arrayrulewidth0.3pt
\caption{Distances $\rho_0$ and $\rho_1$ between texts; see Table I
  and (\ref{1}--\ref{4}). The phoneme frequencies are extracted from
  different words of each text after excluding the words that are
  common for both compared texts; see (\ref{eq:21}, \ref{eq:177}) for
  comparison with the situation without excluding common words.}
\begin{tabular}{||c||c|c|c||c|c|c||c|c|c||}
\hline\hline
   & 12  & 13  & 23   & 45 & 46 & 56 & 78 & 79 & 89    \\
\hline\hline
$\rho_0\times 10^5$ 
&3792 &3217 &3734 &3146 &2930 &2329 &5918 &4421 &4770
\\
\hline
$\rho_1\times 10^5$ 
&2832 &2463 &2502 &2190 &2215 &1610 &3317 &2773 &2809
\\
\hline\hline
\end{tabular}
\begin{tabular}{||c|c|c|c|c|c|c|c|c|c|c|c|c||}
\hline\hline
   & 14 & 15 & 16 & 17 & 18 & 19 & 24 & 25 & 26 & 27 & 28 & 29    \\
\hline\hline
$\rho_0\times 10^5$ 
&4758 &5742 &6087 &12574 &15119 &13490 &5708 &6385 &6880 &13323
&15733 &14113
\\
\hline
$\rho_1\times 10^5$ 
&3912 &4276 &4830 &8800 &9576 &8895 &4529 &4991 &5495 &9469 &10387
&9621
\\
\hline\hline
\end{tabular}
\begin{tabular}{||c|c|c|c|c|c|c|c|c|c|c|c|c|c|c|c||}
\hline\hline
   & 34 & 35 & 36 & 37 & 38 & 39 & 47 & 48 & 49 & 57 & 58 & 59 & 67&
   68& 69    \\
\hline\hline
$\rho_0\times 10^5$ 
&5188  &5887  &6476  &13391 &15842 &14244 &10980 &13905 &12109
&10346 &13003 &11673 &10413 &13288 &11911
\\
\hline
$\rho_1\times 10^5$ 
&4344 &4917 &5285 &9835 &10637 &9891 &7025 &7371 &6928 &6537 &7021
&6673 &6580  &6667 &6433
\\
\hline\hline
\end{tabular}
\end{center}
\end{table*}

\begin{table*}[htbp]
\begin{center}
\tabcolsep0.010in \arrayrulewidth0.3pt
  \caption{The fraction $p$ of common words between texts given in
    Table I. Now $p$ is defined as follows. Let $n(i)$ and $n(ij)$ be,
    respectively, the number of different words in text $i$ and the
    number of common words in texts $i$ and $j$. We define:
    $p(ij)=n(ij)/(n(i)+n(j)-n(ij))$, where $0\leq p(ij)\leq 1$. This
    is the number of common words divided over the number of all
    different words in texts $i$ and $j$. As seen from the data below,
    analogues of (\ref{eq:12}--\ref{eq:14}) hold with $1-p(ij)$
    instead of $\rho_\lambda(ij)$. }
\begin{tabular}{||c||c|c|c||c|c|c||c|c|c||}
\hline\hline
   & 12  & 13  & 23   & 45 & 46 & 56 & 78 & 79 & 89    \\
\hline\hline
$p\times 10^5$ 
&47554 &47786 &50655 &41146 &42454 &41822 &45010 &46948 &48173
\\
\hline\hline
\end{tabular}
\begin{tabular}{||c|c|c|c|c|c|c|c|c|c|c|c|c||}
\hline\hline
   & 14 & 15 & 16 & 17 & 18 & 19 & 24 & 25 & 26 & 27 & 28 & 29    \\
\hline\hline
$p\times 10^5$ 
&35592 &35819 &36660 &28978 &25870 &26730 &32902 &32499 &33877
&26549 &24180 &24643
\\
\hline\hline
\end{tabular}
\begin{tabular}{||c|c|c|c|c|c|c|c|c|c|c|c|c|c|c|c||}
\hline\hline
   & 34 & 35 & 36 & 37 & 38 & 39 & 47 & 48 & 49 & 57 & 58 & 59 & 67&
   68& 69    \\
\hline\hline
$p\times 10^5$ 
&33463 &32813 &34643 &27572 &25340 &25733 &33901 &30387 &32005
&32069 &27963 &29994 &32002 &28649 &30518
\\
\hline\hline
\end{tabular}
\end{center}
\end{table*}

\subsection{The origin of the author-dependency effect is not in
  common words}

One possible reason for the author-dependency of phoneme frequencies
is that the effect is due to the vocabulary of the author. In this
scenario the similarity between phoneme frequencies in text written by
the same author would be caused by the fact that these texts have
sufficiently many common words that carry out the same phonemes.

Texts written by the same author do have a sizeable number of common
words, as was already noted within the authorship attribution research
\cite{joula,ule}. We confirm this result in Table VIII, where it is
seen that the fraction of common words holds the analogues of
(\ref{eq:12}--\ref{eq:14}). Hence this fraction also shows the
author-dependency effect.

In order to understand whether the author-dependency of phoneme
frequencies can be explained via common words, we excluded from
different words of texts $i$ and $k$ the common words of those texts
[$i,k=1,...,9$, see Table I], re-calculated phoneme frequencies, and
only then determined the respective distances $\rho_0^{\rm
  [no~comm.~words]}(ik)$ and $\rho_1^{\rm [no~comm.~words]}(ik)$.  If
the explanation via common words holds, they will not show
author-dependency. This is however not the case: the effect is there
because relations (\ref{eq:12}--\ref{eq:14}) do hold for them
\begin{eqnarray}
  \label{eq:21}
    z_\lambda^{\rm [no~comm.~words]}(a) > 0,
  \qquad \lambda=0,1, \qquad
  a={\rm A, \, D, \, T}.
\end{eqnarray}
Eq.~(\ref{eq:21}) is deduced from Table VII, where we present the
distances $\rho_0$ and $\rho_1$ for the situation, where the common
words are excluded.

After
excluding the common words the author-dependency did not get stronger
in the sense of (\ref{eq:16}), because the data of Tables VI
(different words) and VII (excluded common words) imply for texts
written by the same author
\begin{eqnarray}
  \label{eq:177}
&&\rho^{\rm [no~comm.~words]}_\lambda(ij) > \rho^{\rm
[diff.~words]}_\lambda(ij),\qquad  \lambda=0,1, \\
&& \quad
(ij)=\{\, (12), (13), (23), (45), (46), (78), (79), (89)\,  \}.  \nonumber
\end{eqnarray}
In this context recall (\ref{eq:20}, \ref{eq:16}). But it also did not
get weaker [cf.~(\ref{eq:20}) and (\ref{eq:12}--\ref{eq:14})], because
\begin{eqnarray}
  \label{eq:17}
   z_\lambda^{\rm [no~comm.~words]}(a) > z_\lambda^{\rm [diff.~words]} (a),
  \qquad \lambda=0,1, \qquad
  a={\rm A, \, D, \, T},
\end{eqnarray}
as seen from Tables VI and VII, which refer, respectively, to
different words and the excluded common words.

\comment{

The qualitative explanation of () can be as follows. It is known that
the words in a text can separated into functional words, content words
that are not specific for the text, key-words (i.e.  content words
that are specific for the text) and rare words (they are met few times
in text, but overall amount to a sizeable fraction of the set of
different words).

The latter two groups are specific for a given text and cannot be a
part of a text-independent author's vocabulary. The first group is
basically the same for all authors, hence it also not involved in the
author's vocabulary. So the definition of the author's vocabulary can
be based only on content words that are not specific for the
text. Such content words can display essential author-dependency, but
it is not clear whether they can explain the author-dependency of
phoneme frequencies.

Summarizing this discussion, we are left with the option that the
uncovered author-dependency effect for phoneme frequencies relates to
storage (by the authors) of structures lower than the word: syllables
and/or phonemes.  }

\section{Summary}

Phonemes are the minimal building blocks of the linguistic hierarchy
that still relate to meaning.  A coarse-grained description of phoneme
frequencies is provided by rank-frequency relations. For describing
these relations we followed the qualitative analogy between atoms and
phonemes \cite{zwick,abler}. Atoms amount to a finite (and not very
large) number of discrete elements from which the multitude of
substances and materials are built \cite{balian}. Likewise, a finite
number of phonemes can construct a huge number of texts \cite{abler}.

The simplest description of an (sufficiently dilute) atomic system is
provided via the ideal gas model \cite{balian}.  By studying 16
native-English authors, we show that the rank-frequency relations for
phonemes can be described via the ordered statistics of the Dirichlet
density, the direct analogue of the ideal gas model in statistics.  In
particular, though the number of phonemes is not very large (English
has 44 phonemes), it is just large enough to validate the statistical
description. The single parameter of the Dirichlet density corresponds
to the (inverse) temperature of the ideal gas in statistical
physics. It appears that the most frequent phonemes fit the Dirichlet
distribution much better than others. This contrasts to the
rank-frequency relations for words, where the Zipf's law holds worst
for the most frequent words.

The fitting to the Dirichlet density uncovers an important aspect of
phoneme frequencies: they depend on the author of the text.  This fact
is seen for authors who created their works in various genres (novels,
scientific texts, journal papers), and also for authors whose
language-dependence on social, temporal and educational background has
been minimized (e.g. the closely inter-related group of English
naturalists including Darwin, Wallace, Lyell, and Spencer). We
confirmed this result via a parameter-free method that is based on
calculating distances between phoneme frequencies of different texts.
Again, this contrasts to the Zipf's law for rank-frequency relations of
words whose shape is author-independent.

It is well-known that certain aspects of text-statistics display
author-dependency, and this is applied in various author attribution
tasks; see e.g. \cite{gibbs,ule,joula,koppel,stama,kuku} for recent
reviews. In particular, this concerns frequencies of functional
words. The fact that author-dependency is seen on such a
coarse-grained level as rank-frequency relations may mean that phoneme
frequencies can be useful for existing methods of authorship
attribution \cite{joula,koppel,stama,kuku}. This should be clarified
in future.

A straightforward reason for explaining the author-dependency effect of
phoneme frequencies would be that it is due to the author's
vocabulary, as reflected by common words in texts written by the same
author. The previous section has shown that such an explanation is
ruled out.

Then we are left with options that the effect is due to
storing (with different frequencies) syllables or/and phonemes. If
syllable frequencies have author-dependency, this could result to
author-dependent phoneme frequencies, because there are specific rules
that (at least probabilistically) determine the phoneme composition of
syllables \cite{kessler}. But note that syllables are in several respect
similar to words (and not phonemes): {\it (i)} there are many of them;
e.g. English has more than 12000 syllables. {\it (ii)} There is large
gap between frequent and infrequent syllables \cite{levelt} (cf. with
the hapax legomena for words). {\it (iii)} There are indications that
syllables are stored in a syllabic lexicon that in several ways is similar to
the mental lexicon that stores words \cite{levelt}.

The second possibility would mean that the authors store phonemes
\cite{nathan}, and this will provide a statistical argument for
psychological reality of phonemes. Note that the issue of
psychological reality of a phoneme is not settled in modern phonology
and psychology, various schools arguing pro and contra of it; see
\cite{port,suomi,nathan,savin,foss} for discussions. And then both
these options might be present together. Thus further research|also
involving rank-frequency relations for syllables|is needed for
clarifying the situation.

The presented methods can find applications in animal communication
systems. In this context, we recall an interesting argument
\cite{ivanov}. The number of phonemes in languages roughly varies
between $20$ and $50$. Indeed, the average number of phonemes in
European languages is $\simeq 37$. (English has 44 phonemes, but if
diphthongs are regarded as combinations of a vowel and a semi-vowel
this number reduces to 36.)  In tonal languages the overall number of
phonemes is larger, e.g. it is $\sim 180$ for Chinese. (The tone
produces phonemes and not allophones, since they do change the
meaning.) But the number of phonemes without tones still complies with
the above rough bound. Since Old Chinese (spoken in 11 to 7'th
centuries B.C.) lacked tones, the tonal phonemes of modern Chinese
evolved from their non-tonal analogues that complies with the above
number \cite{sampson}.  By its order of magnitude this number ($\sim
20-50$) coincides \cite{ivanov} with the number of ritualized
(i.e. sufficiently abstract) signals of animal communication, which is
also stable across different species \cite{moynihan}. (An example of
this are gestures of apes.) This number is sufficiently large to
invite the application of the presented statistical methods to signals
of animal communication. And the stability of this number may mean
that there are further similarities (to be yet uncovered) between
phonemes and ritualized signals.

\section*{Acknowledgements}

This work was supported by National Natural Science Foundation of
China (Grant No. 11505071), the Programme of Introducing Talents of
Discipline to Universities under Grant NO. B08033.

\clearpage

\section*{\large Supplementary material}

\comment{
\section{Some remarks about the number of phonemes}
\label{phopho}

The number of phonemes, syllables, morphemes and words have roughly
the same order of magnitude for different languages: the number of
phonemes varies between $10$ and $\sim 50$, there are $10^2-10^3$
syllables, few thousands morphemes, $10^4-10^5$ words and practically
unlimited amount of texts.

Note that English has 44 phonemes. But if diphthongs are regarded as
combinations of a vowel and a semi-vowel (and not as separate phonemes)
this number reduces to 36. The average number of phonemes in European
languages is $\simeq 37$.

Now in tonal languages the overall number of phonemes is larger,
e.g. it is $\sim 180$ for Chinese. (The tone produce phonemes and not
allophones, since they do change the meaning.) But the number of
phonemes without tones still complies with the above rough
bound. Since Old Chinese (spoken in 11 to 7'th centuries B.C.) lacked
tones, the tonal phonemes of modern Chinese evolved from their
non-tonal analogues \cite{sampson}.

Another interesting point about the above rough number of phonemes is
that by its order of magnitude this number coincides \cite{ivanov}
with the number of ritualized (i.e. sufficiently abstract) signals of
animal communication, which is also stable across different species
\cite{moynihan}. An example of this are gestures for monkeys.
}

\section*{I. English phonemes}

\label{english}

Here we recall $44$ English phonemes according to the International
Phonetic Alphabet.

\noindent I. $20$ vowels (7 short phonemes, 5 long and 8 diphtongs):

\textturnv \, (b\underline{u}t),
\ae \, (c\underline{a}t), \textschwa\, (\underline{a}bout), e
(m\underline{e}n),  \i \, (s\underline{i}t),
\textturnscripta \, (n\underline{o}t),
\textupsilon \, (b\underline{oo}k),

\textscripta : (p\underline{ar}t),
\textrevepsilon : (w\underline{or}d,
l\underline{ear}n),
i: (r\underline{ea}d),
\textopeno : (s\underline{o}rt),
u: (t\underline{oo}),

a\i\, (m\underline{y}), a\textupsilon\, (h\underline{ow}),
o\textupsilon\, (g\underline{o}),
e\i\, (d\underline{ay}), \i\textschwa\, (h\underline{ere}), o\i \,
(b\underline{oy}), \textupsilon \textschwa\, (t\underline{our},
p\underline{ur}e), e\textschwa \, (w\underline{ear}, f\underline{air})

\noindent II. $24$ consonants:

 b \, (\underline{b}orn), d \, (\underline{d}o), f \,
  (\underline{f}ive), g \, (\underline{g}et), h \,
  (\underline{h}ouse), j \, (\underline{ye}s), k \, (\underline{c}at),
  l \, (\underline{l}ion), m \, (\underline{m}ouse), n \,
  (\underline{n}ouse), \textrtailn \, (si\underline{ng}), p \,
  (\underline{p}ut), r \, (\underline{r}oom), s \, (\underline{s}aw),
  \textesh \, (\underline{sh}all), t \, (\underline{t}ime),
  \textteshlig \, (\underline{ch}ur\underline{ch}), \texttheta \,
  (\underline{th}ink), $\eth$ \, (\underline{th}e), v \,
  (\underline{v}ery), w \, (\underline{w}indow), z \,
  (\underline{z}oo), \textyogh \, (ca\underline{s}ual), d\textyogh \,
  (\underline{j}u\underline{dg}e)

\comment{
\subsection{Mandarin Chinese phonemes}

In Mandarin Chinese, each character corresponds to one syllable,
the pronunciation of which is composed by three elements: initial
sound, final sound and tone. The initial sounds are consonants and
the final sounds contain at least one vowel. Some syllables
consist only an initial sound or a final sound.

\noindent I. There are 23 initial sounds:

{\bf b, p, m, f, d, t, n, l, g, k, h, j, q, x, z, c, s, zh, ch,
sh, r, y, w}

\noindent II. There are 36 final sounds:

6 simple finals: {\bf a, o, e, i, u, \"{u} }

14 compound finals: {\bf ai, ao, ei, ia, iao, ie, iou, ou, ua,
uai, \"{u}e, uei, uo, er}

16 nasal finals:

8 front nasals: {\bf an, en, ian, in, uan, \"{u}an, uen, \"{u}n}

8 back nasals: {\bf ang, eng, ing, ong, iang, iong, uang, ueng}

\noindent III. For each final sound, it has five different tonal
pronunciations, for example,

{\bf \=a, \'a, \v a, \`a}; {\bf \=an, \'an, \v an, \`an}; {\bf
\=ang, \'ang, \v ang, \`ang} ...

}

\section*{II. Order statistics for Dirichlet density}

\label{orde}

Let us introduce the following notation for the order integration
\begin{eqnarray}
  \label{eq:a1}
    {\cal I}(\d \theta_1,...,\d \theta_n)
\equiv \int_0^\infty
  \d \theta_1
  \int_0^{\theta_1}
  \d \theta_2 ...
  \int_0^{\theta_{n-1}}
  \d \theta_n .
\end{eqnarray}

Now the average over the order statistics of the Dirichlet density is
defined as
\begin{eqnarray}
  \label{eq:a2}
  \langle \theta_{(r)}^m\rangle =\frac{    {\cal I}(\d \theta_1,...,\d
    \theta_n)\, \theta_r^m \,\delta(\sum_{k=1}^n \theta_k-1)\,
\prod_{k=1}^n \theta_k^{\beta-1} }
{    {\cal I}(\d \theta_1,...,\d \theta_n)\, \delta(\sum_{k=1}^n
  \theta_k-1)\,
\prod_{k=1}^n \theta_k^{\beta-1}}.
\end{eqnarray}
In the numerator of (\ref{eq:a2}) we change variables as
$\hat{\theta}_k=r\theta_k$ ($r>0$), multiply both sides by $e^{-r}$,
and then integrate both sides over $r\in [0,\infty)$:
\begin{eqnarray}
  \label{eq:a3}
  {\cal I}(\d \theta_1,...,\d
    \theta_n)\, \theta_r^m \,\delta(\sum_{k=1}^n \theta_k-1)\,
\prod_{k=1}^n \theta_k^{\beta-1} \times \Gamma[n\beta+m]
=  {\cal I}(\d \hat{\theta}_1,...,\d
    \hat{\theta}_n)\, \hat{\theta}_r^m
\prod_{k=1}^n \hat{\theta}_k^{\beta-1}e^{-\hat{\theta}_k}.
\end{eqnarray}
The denominator of (\ref{eq:a2}) is worked out analogously.

Let us now define
\begin{eqnarray}
  \label{eq:a4}
  \chi_r(y;m)=\frac{\Gamma[n\beta]}{\Gamma[n\beta+m]}\,
\frac{{\cal I}(\d \hat{\theta}_1,...,\d
    \hat{\theta}_n)\, \delta(y-\hat{\theta}_r)\,
\prod_{k=1}^n \hat{\theta}_k^{\beta-1}e^{-\hat{\theta}_k}}
{{\cal I}(\d \hat{\theta}_1,...,\d
    \hat{\theta}_n)\,
\prod_{k=1}^n \hat{\theta}_k^{\beta-1}e^{-\hat{\theta}_k}}
\end{eqnarray}
so that the following relation holds
\begin{eqnarray}
  \label{eq:aa5}
\langle \theta_{(r)}^m\rangle =\int_0^\infty \d y\, y^m   \chi_r(y;m).
\end{eqnarray}
This is the equation (9) of the main text. Working out ${\cal I}(\d
\hat{\theta}_1,...,\d \hat{\theta}_n)\, \prod_{k=1}^n
\hat{\theta}_k^{\beta-1}e^{-\hat{\theta}_k}$ and ${\cal I}(\d
\hat{\theta}_1,...,\d \hat{\theta}_n)\, \delta(y-\hat{\theta}_r)\,
\prod_{k=1}^n \hat{\theta}_k^{\beta-1}e^{-\hat{\theta}_k}$ in
(\ref{eq:a4}) via integration by parts (starting from the last
integration in ${\cal I}(\d \hat{\theta}_1,...,\d \hat{\theta}_n)$) we
obtain equations (7--9) of the main text.

If $(n-r)\gg 1$ and $r\gg 1$ the behavior of $\chi_r(y;m)$ in
equations (7) of the main text
is determined by the exponential factor
$e^{(n-r)\ln\varphi(y)+r\ln (1-\varphi(y))}$. Working it out via the
saddle-point method we conclude that asymptotically:
\begin{eqnarray}
  \label{eq:a8}
      \chi_r(y;m)\simeq\frac{\Gamma[n\beta]}{\Gamma[n\beta+m]}\,
\frac{1}{\sqrt{2\pi\sigma}}\,
e^{-\frac{1}{2\sigma}(y-y_0)^2},
\end{eqnarray}
where $y_0$ and $\sigma$ are defined as follows
\begin{eqnarray}
  \label{eq:a9}
  \frac{n-r}{n}=\varphi(y_0), ~~~~ \sigma=\frac{(n-r)r}{n^3}\,
\frac{1}{[\,\varphi'(y_0)\,]^2},
\end{eqnarray}
where $\phi'(y)=\d \varphi(y)/\d y$.

Hence we get from (\ref{eq:aa5}) and (\ref{eq:a8}, \ref{eq:a9}):
\begin{eqnarray}
  \label{eq:a10}
  &&  \langle \theta_{(r)}\rangle=\frac{y_0}{n\beta},\\
  &&\langle \theta_{(r)}^2\rangle- \langle \theta_{(r)}\rangle^2 =
  \frac{n\beta\sigma-y_0^2}{[n\beta]^2(n\beta+1)}
= \frac{1}{[n\beta]^2(n\beta+1)}\left
    (\frac{\beta(n-r)r}
    {n^2}\,\frac{1}{[\varphi'(y_0)]^2}-y_0^2\right).
  \label{eq:a11}
\end{eqnarray}

The importance of fluctuations is characterized by
\begin{eqnarray}
  \label{eq:a12}
  \frac{\langle \theta_{(r)}^2\rangle- \langle \theta_{(r)}\rangle^2}
{\langle \theta_{(r)}\rangle^2}&=&
\frac{1}{n\beta+1}\left
(\frac{\beta(n-r)r}
{n^2}\,\frac{1}{y_0^2[\varphi'(y_0)]^2}-1\right)\\
&=& \frac{1}{n\beta+1}\left(
\frac{\beta(n-r)r}{n^2}\,\,
\Gamma^2[\beta]\,y_0^{-2\beta}\,e^{2y_0}
-1\right),
  \label{eq:a13}
\end{eqnarray}
where we employed
\begin{eqnarray}
  \label{eq:aa7}
  \varphi (y)=\frac{1}{\Gamma[\beta]}\int_0^y\d x\, x^{\beta-1}e^{-x}
\end{eqnarray}
This is the equation (8) of the main text. Eq.~(\ref{eq:a13}) is a
good approximation of $\frac{\langle \theta_{(r)}^2\rangle- \langle
  \theta_{(r)}\rangle^2} {\langle \theta_{(r)}\rangle^2}$ calculated
(exactly) from equations (7-9) of the main text; see Fig.~1 of the
main text.

\section*{III. Information on the other 13 authors and 39 texts}

Here are the works by the other 13 English writers we studied in addition to the
authors described in Table I of the main text. After the title of each
work we give its writing/publication date, the number of different
words, and the number of phonemes of different words. The table below
summarizes the values of $\beta$ for phonemes extracted from different
words of each text.

Charlotte Bronte:
{\it Jane Eyre} (1847, 12488, 75933),
{\it Shirley} (1849, 14481, 88911),
{\it Villette} (1853, 14176, 88025).

Clive S. Lewis: {\it Perelandra} (1943, 7030, 41265),
{\it Out of the Silent Planet} (1938, 6045, 35371),
{\it That Hideous Strength} (1946, 7842, 46618).

George Eliot: {\it Adam Bede} (1859, 9685, 56819),
{\it Romola} (1862, 13402, 83255),
{\it The Mill on the Floss} (1860, 11682, 72071).

George MacDonald: {\it Paul Faber, Surgeon} (1879, 9615, 57634), {\it
  There and Back} (1891, 8807, 51865), {\it Unspoken Sermons, Series
  I-III} (1867--1889), 7815, 47674).

Alfred R. Wallace: {\it Contributions to the Theory of Natural
Selection} (1870, 6829, 43619),
{\it Man's Place in the Universe} (1904, 5626, 35738),
{\it The Malay Archipelago} (1869, 8785, 52760).

Charles Darwin: {\it On the Origin of Species} (1859, 6764, 42519),
{\it The Descent of Man, and Selection in Relation to Sex} (1871,
13069, 82027), {\it The Voyage of the Beagle} (1839, 11359, 69667).

Herbert G. Wells: {\it Marriage} (1912, 12076, 76705), {\it The
  Country of the Blind, and Other Stories} (1894-1909, 11537, 71091),
{\it The New Machiavelli} (1911, 12702, 81773).

Herbert Spenser:
{\it The principle of psychology} (1855, 6932, 48036),
{\it The Principles of Ethics} (1897, 10575, 71729),
{\it The Principles of Sociology} (1874, 15215, 98353).

Joseph R. Kipling:
{\it A Diversity of Creatures} (1912, 9993, 57358),
{\it From Sea to Sea; Letters of Travel} (1889, 15165, 91038),
{\it Indian Tales} (1890, 11975, 69231).

Oscar Wilde: {\it A Critic in Pall Mall Being Extracted from Reviews
  and Miscellanies} (1919, 8168, 49299), {\it Miscellanies} (1908,
8204, 50539), {\it The Picture of Dorian Gray} (1891, 6725, 37933).

Charles Lyell:
{\it A Manual of Elementary Geology} (1852, 9573, 61983),
{\it The Antiquity of Man} (1863, 9280, 59868),
{\it The Student's Elements of Geology} (1865, 10347, 67742).

Walter Scott:
{\it Ivanhoe, A Romance} (1819, 11857, 71974),
{\it Old Mortality} (1816, 12049, 73894),
{\it Rob Roy} (1817, 12524, 76175).

William M. Thackeray:
{\it The History of Pendennis} (1848, 15591, 96039).
{\it The Virginians} (1857, 15158, 92548).
{\it Vanity Fair} (1848, 14695, 90373).

\begin{table*}[htbp]
\begin{center}
\tabcolsep0.060in \arrayrulewidth0.3pt
\caption{The values of $\beta$ extracted from different words of each
  text. The order of texts corresponds to the description, i.e. the
  three texts by C. Bronte in the table (from left to right) refer,
  respectively, to {\it Jane Eyre}, {\it Shirley}, and {\it Villette}.
}
\begin{tabular}{||c||c|c|c||}
\hline\hline
C. Bronte
& 0.762   & 0.767  & 0.758  \\
\hline\hline
C. S. Lewis
& 0.781  & 0.780  & 0.778  \\
\hline\hline
G. Eliot
& 0.747  & 0.741  & 0.748  \\
\hline\hline
G. MacDonald
& 0.773  & 0.773  & 0.766  \\
\hline\hline
A. R. Wallace
& 0.744  & 0.756  & 0.739  \\
\hline\hline
C. Darwin
& 0.817  & 0.810  & 0.822  \\
\hline\hline
H. G. Wells
& 0.737  & 0.735  & 0.724  \\
\hline\hline
H. Spenser
& 0.646   & 0.658  & 0.650  \\
\hline\hline
J. R. Kipling
&0.868   & 0.852  & 0.872  \\
\hline\hline
O. Wilde
& 0.793  & 0.785  & 0.803  \\
\hline\hline
C. Lyell
& 0.798  & 0.785  & 0.792  \\
\hline\hline
W. Scott
& 0.808  & 0.795  & 0.787  \\
\hline\hline
W. M. Thackeray
& 0.818  & 0.815  & 0.818 \\
\hline\hline
\end{tabular}
\end{center}
\end{table*}

The table shows that the values of $\beta$ for several authors do
overlap. These overlaps are accidental, as can be verified by
calculating the distances. Here are some examples for authors whose
values of $\beta$ overlap:
\begin{gather}
  \label{eq:19}
  {\rm max}_{\rm \, Darwin}[\rho_0]= 0.001155, \qquad
  {\rm max}_{\rm \, Thackeray }[\rho_0]=0.0097, \\
  {\rm min}_{\rm \, Darwin\, vs. \, Thackeray }[\rho_0]=0.01508,
\end{gather}
where ${\rm max}_{\rm \, Darwin}[\rho_0]$ is the maximal
$\rho_0$-distance between the 3 texts by Darwin [see (20) of the main
text for the definition of $\rho_0$], ${\rm max}_{\rm \, Thackeray
}[\rho_0]$ is the same quantity for the texts by Thackeray, and $ {\rm
  min}_{\rm \, Darwin\, vs. \, Thackeray }[\rho_0]$ is the minimal
$\rho_0$-distance between the texts of Darwin versus those of
Thackeray. It is seen that although the values of $\beta$ for Darwin
and Thackeray overlap, the distances between phoneme frequencies do
cluster, and they hold analogues of (21--23) of the main text,
i.e. ${\rm max}_{\rm \, Darwin}[\rho_0] < {\rm min}_{\rm \, Darwin\,
  vs. \, Thackeray }[\rho_0]$ and $ {\rm max}_{\rm \, Thackeray
}[\rho_0]< {\rm min}_{\rm \, Darwin\, vs. \, Thackeray }[\rho_0]$.

Similar relations hold for the $\rho_1$ distance [see (19) of the main
text]:
\begin{gather}
  {\rm max}_{\rm \, Darwin}[\rho_1]= 0.01674, \qquad
  {\rm max}_{\rm \, Thackeray }[\rho_1]=0.00943, \\
  {\rm min}_{\rm \, Darwin\, vs. \, Thackeray }[\rho_1]=0.01705,
\end{gather}

We give several other examples of distances for those authors whose
values of $\beta$ overlap. We found that all these examples hold the
above clustering feature.

\begin{gather}
  {\rm max}_{\rm \, Lyell}[\rho_0]= 0.03976, \qquad
  {\rm max}_{\rm \, MacDonald }[\rho_0]=0.02557, \\
  {\rm min}_{\rm \, Lyell\, vs. \, MacDonald }[\rho_0]=0.04632, \\
 {\rm max}_{\rm \, Lyell}[\rho_1]= 0.02442, \qquad
  {\rm max}_{\rm \, MacDonald }[\rho_1]=0.01015, \\
  {\rm min}_{\rm \, Lyell\, vs. \, MacDonald }[\rho_1]=0.02968.
\end{gather}
\begin{gather}
 {\rm max}_{\rm \, Wallace}[\rho_0]= 0.02508, \qquad
  {\rm max}_{\rm \, Eliot }[\rho_0]=0.02483, \\
  {\rm min}_{\rm \, Wallace\, vs. \, Eliot }[\rho_0]=0.02821, \\
  {\rm max}_{\rm \, Wallace}[\rho_1]= 0.0135, \qquad
  {\rm max}_{\rm \, Eliot }[\rho_1]=0.01108, \\
  {\rm min}_{\rm \, Wallace\, vs. \, Eliot }[\rho_1]=0.02047.
\end{gather}
\begin{gather}
  {\rm max}_{\rm \, Wilde}[\rho_0]= 0.0161, \qquad
  {\rm max}_{\rm \, Scott }[\rho_0]=0.01801, \\
  {\rm min}_{\rm \, Wilde\, vs. \, Scott }[\rho_0]=0.02851, \\
  {\rm max}_{\rm \, Wilde}[\rho_1]= 0.01901, \qquad
  {\rm max}_{\rm \, Scott }[\rho_1]=0.01455, \\
  {\rm min}_{\rm \, Wilde\, vs. \, Scott }[\rho_1]=0.01954.
\end{gather}

\end{document}